\documentclass[runningheads]{llncs}
\usepackage{graphicx}

\usepackage{tikz}
\usepackage{comment}
\usepackage{amsmath,amssymb} 
\usepackage{color}

\usepackage{pifont}
\usepackage{caption}
\usepackage{multirow}
\usepackage{subcaption}
\usepackage{wrapfig}
\usepackage{boldline}

\usepackage{float}
\usepackage{algorithm}
\usepackage{algpseudocode}
\usepackage[frozencache,cachedir=.]{minted}
\usepackage[pagebackref=true,breaklinks=true,letterpaper=true,colorlinks,bookmarks=false]{hyperref}
\usepackage{comment}

\newcommand{\cmark}{\ding{51}}

\begin{document}
\pagestyle{headings}
\mainmatter
\def\ECCVSubNumber{3169}  

\title{Fast-MoCo: Boost Momentum-based Contrastive Learning with Combinatorial Patches}

\titlerunning{Fast-MoCo: Boost Momentum-based Contrastive Learning}
%
\author{Yuanzheng Ci\inst{1} \and
Chen Lin\inst{2} \and
Lei Bai\inst{3}\thanks{Corresponding author} \and
Wanli Ouyang\inst{3,1}}
\authorrunning{Y. Ci et al.}
%
\institute{
The University of Sydney, SenseTime Computer Vision Group\\
\email{\{yuanzheng.ci,wanli.ouyang\}@sydney.edu.au} \and
University of Oxford \\ 
\email{chen.lin@eng.ox.ac.uk} \and
Shanghai AI Laboratory \\
\email{bailei@pjlab.org.cn}
}
\maketitle

\begin{abstract}
Contrastive-based self-supervised learning methods achieved great success in recent years. However, self-supervision requires extremely long training epochs (e.g., 800 epochs for MoCo v3) to achieve promising results, which is unacceptable for the general academic community and hinders the development of this topic. This work revisits the momentum-based contrastive learning frameworks and identifies the inefficiency in which two augmented views generate only one positive pair.
We propose Fast-MoCo - a novel framework that utilizes combinatorial patches to construct multiple positive pairs from two augmented views, which provides abundant supervision signals that bring significant acceleration with neglectable extra computational cost. Fast-MoCo trained with \textbf{100} epochs achieves $\bf{73.5\%}$ linear evaluation accuracy, similar to MoCo v3 (ResNet-50 backbone) trained with 800 epochs. Extra training (\textbf{200} epochs) further improves the result to $\bf{75.1\%}$, which is on par with state-of-the-art methods. Experiments on several downstream tasks also confirm the effectiveness of Fast-MoCo.\footnote{Code and pretrained models are available at 
\href{https://github.com/orashi/Fast-MoCo}{https://github.com/orashi/Fast-MoCo}}  
\end{abstract}

\section{Introduction}

\begin{figure*}[t]
    \centering
    \begin{subfigure}[b]{0.38\textwidth}
        \centering
        \includegraphics[width=\textwidth]{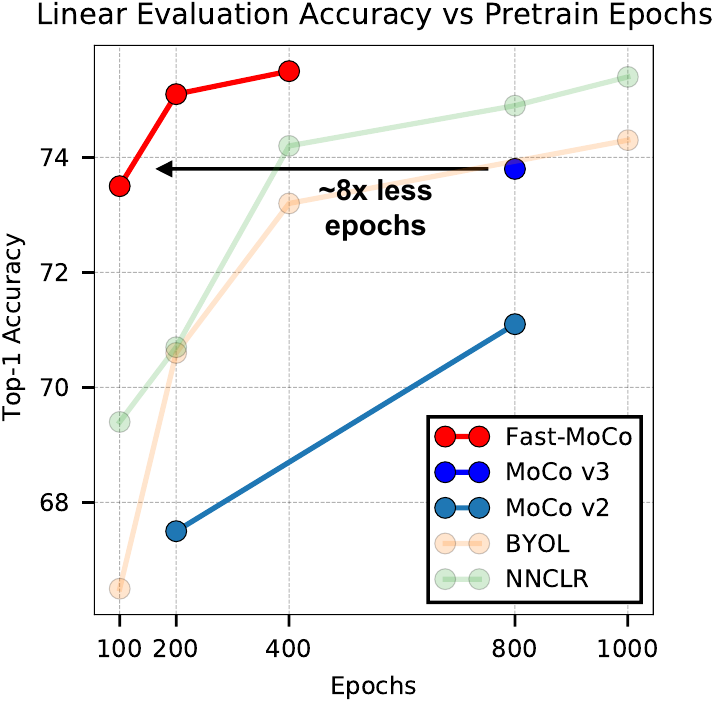}
        \caption{}
    \end{subfigure}
    \hfill
    \begin{subfigure}[b]{0.6\textwidth}
        \centering
        \includegraphics[width=\textwidth]{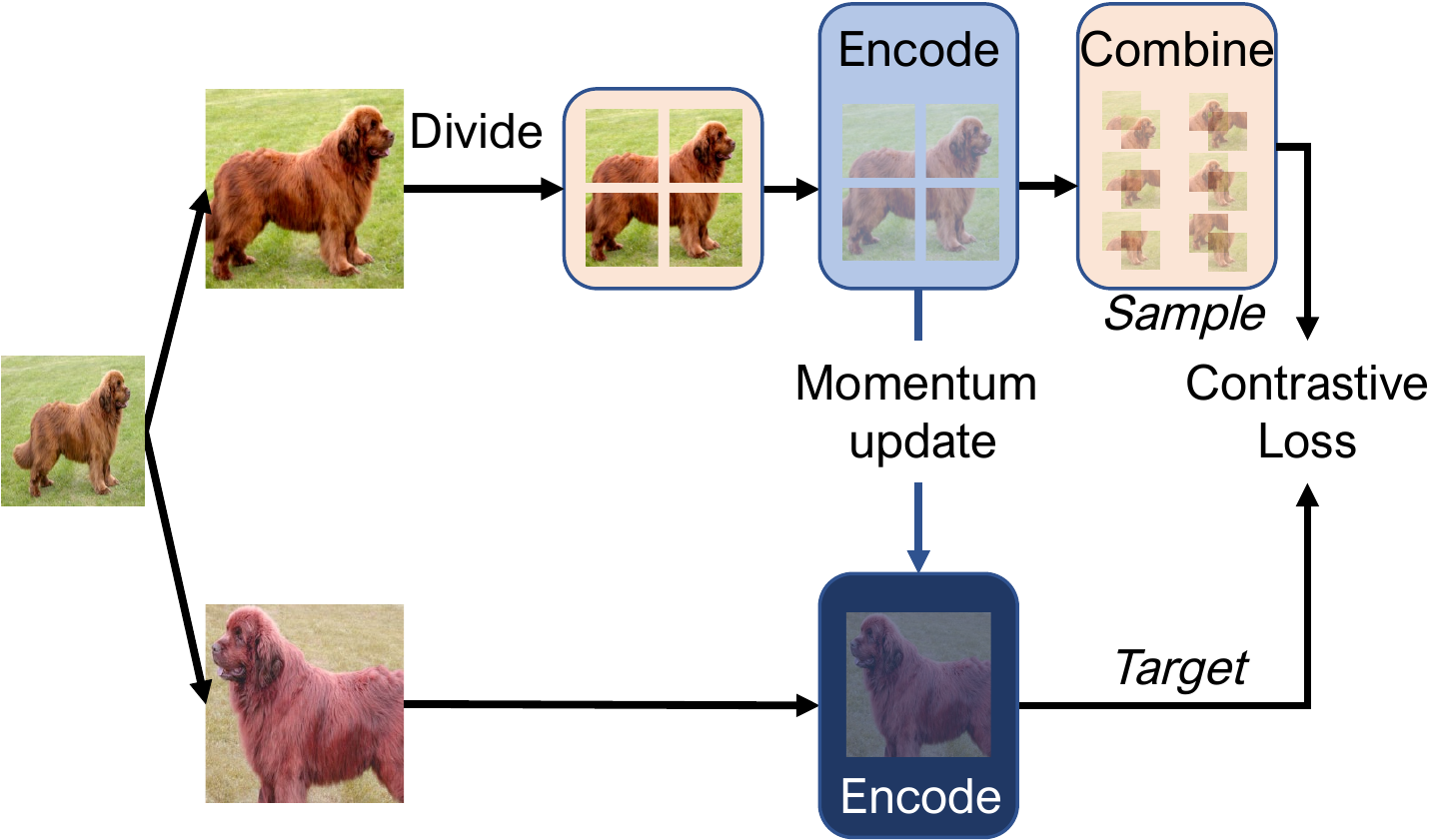}
        
        \caption{}
    \end{subfigure}
    \caption{(a): Comparison with state-of-the-arts on ImageNet. All methods uses ResNet-50 encoders and are measured with Top-1 linear evaluation accuracy. (b): Overview of Fast-MoCo that includes the Split-Encode-Combine pipeline.}
    \label{fig:intro}
\end{figure*}

Self-supervision is crucial in some of the most remarkable achievements from natural language processing (NLP)~\cite{devlin2018bert,brown2020language} to computer vision \cite{chen2020simple}. In particular, recent advances in contrastive learning produced state-of-the-art results on self-supervised learning benchmarks~\cite{grill2020bootstrap,chen2021empirical,wang2022revisiting}. Contrastive learning performs an instance discrimination pretext task by attracting the embedding of positive samples closer while encouraging the negative samples to be further apart. Some methods opt to make the sample pairs asymmetric with tools such as momentum encoder~\cite{he2020momentum}, predictor~\cite{grill2020bootstrap} and \texttt{stop-grad}~\cite{chen2021exploring} to provide more flexibility for architecture design~\cite{grill2020bootstrap,gidaris2021obow}.

While great advances have been achieved in the self-supervised learning area in the past two years, a major concern about these works is the extremely long training steps to get a promising performance(e.g., normally 800 epochs, and even 1000 epochs for some methods~\cite{chen2021empirical,grill2020bootstrap,zbontar2021barlow,dwibedi2021little}), which makes it hard or even impossible for many academics to contribute to this area. High training cost also posts challenges when dealing with large industry scale datasets \cite{bao2021beit,he2021masked}. In order to accelerate training, we spotted one limitation of recent momentum based contrastive learning methods~\cite{he2020momentum,chen2020improved,grill2020bootstrap}, which is the \emph{two-image-one-pair} strategy. In this strategy, two images (or two augmented views of the same image) are fed to the deep models separately and then used as one pair for contrastive learning in~\cite{he2020momentum,chen2020improved,chen2021empirical,dwibedi2021little}. 
Although symmetric loss designs are normally employed to improve the sample efficiency, we argue that the \emph{two-image-one-pair} mechanism is sub-optimal.
To overcome this issue, we propose combinatorial patches, a novel mechanism to efficiently generate feature embeddings for arbitrary combination of local patches. In this strategy, an image pair can be used for generating multiple positive pairs for contrastive learning. Therefore, in contrast to the  \emph{two-image-one-pair} mechanism in existing works, our combinatorial patches enable the \emph{two-image-multi-pair} mechanism.  With more pairs used for contrastive learning using this  \emph{two-image-multi-pair} mechanism, our Fast-MoCo method trained using 100 epochs based on MoCo v3 (\emph{two-image-one-pair} mechanism) for ResNet50 can achieve on-par accuracy when compared with MoCo v3 trained using 800 epochs, as shown in Fig.~\ref{fig:intro}(a).

To implement the  \emph{two-image-multi-pair} mechanism, this paper  proposes the Divide-Encode-Combine and then Contrast pipeline as shown in Fig.~\ref{fig:intro}(b). In detail, we divide the input into multiple local patches without overlap in the data preparation stage and encode the local patches by  deep models separately, then combine the encoded features of multiple patches before computing the contrastive loss. We validate various strategies and hyperparameters for both divide and combine stages and provided a detailed analysis across different settings.

We evaluate our method on ImageNet with the ResNet-50 backbone. In a linear evaluation setting, our method achieves $73.5\%$ with only 100 epochs of SSL pretraining, which is $8\times$ faster than the original MoCo to achieve comparable performance. A longer training (400 epochs) further boosts the performance from $73.5\%$ to $75.5\%$. We also tested the learned embeddings in semi-supervised learning, object detection, and instance segmentation. Our method performs better than previous approaches in both settings, which suggests the embeddings learned with our method are general and transferable.

\section{Related Works}

\subsection{Patch Based Representation Learning}
Various self-supervised learning methods~\cite{noroozi2016unsupervised,van2018representation,henaff2020data,gidaris2021obow,chen2021jigsaw,pathak2016context,bao2021beit,he2021masked} manipulates image patches. 
A common way to incorporate patches is to encode them separately~\cite{noroozi2016unsupervised,van2018representation,henaff2020data,gidaris2021obow}, while Jigsaw Clustering~\cite{chen2021jigsaw} encodes multiple patches at the same time: patches are augmented independently and stitched to form a new image for encoding, the encoded features are then separated spatially before pooling to get the embedding for each patch. Either way, the encoded embeddings can then be used for solving jigsaw puzzles~\cite{noroozi2016unsupervised,chen2021jigsaw}, contrastive prediction~\cite{van2018representation,henaff2020data,chen2021jigsaw} or bag-of-word reconstruction~\cite{gidaris2021obow}.
On the other hand, Context encoder~\cite{pathak2016context} encodes an image with random masking and then learns to reconstruct the missing part with a decoder. With a ViT encoder, BEiT~\cite{bao2021beit} and MAE~\cite{he2021masked} split the image into a grid of patches and mask out some of them, the rest patches are gathered and forwarded to get encoded embeddings. They are then optimized for reconstructing the missing patches at feature-level~\cite{bao2021beit} or pixel-level~\cite{he2021masked}. However, these methods do not construct multiple pairs of samples from combinatorial patches and thus are different from our Divide-Encode-Combine pipeline.

\subsection{Contrastive Learning}
Contrastive learning methods~\cite{hadsell2006dimensionality,chen2020simple,caron2020unsupervised} have attracted many attentions for their simplicity and performance. They retrieve useful representations by promoting instance discrimination, where the positive samples are generated by applying different data augmentations to the same image while having an identical spatial size. SwAV~\cite{caron2020unsupervised} and NNCLR~\cite{dwibedi2021little} further extend the semantic gap between a positive pair with a target embedding being replaced by a learned cluster center and a neighborhood embedding. Since the methods in~\cite{hadsell2006dimensionality,chen2020simple,caron2020unsupervised,dwibedi2021little} are not momentum-based learning, our method does not aim at improving them. Besides, our proposed Divide-Encode-Combine scheme is not investigated in them. 

Momentum-based contrastive learning methods adopt an asymmetric forward path. On the online path, an input image is fed into the encoder. On the target path, another input image is fed into a slowly moving momentum encoder~\cite{he2020momentum,chen2020improved,chen2021empirical}. The two encoded samples from these two paths form a pair for contrastive learning, which has been proven to be effective in many scenarios~\cite{gidaris2021obow,grill2020bootstrap,caron2021emerging}. However, these works adopt the \emph{two-image-one-pair} mechanism. In contrast, our Fast-MoCo adopts a \emph{two-image-multi-pair} mechanism. At almost the same training cost of the \emph{two-image-one-pair} mechanism, Fast-MoCo generates more sample pairs in a mini-batch for efficiency.

\section{Method}
In this Section, we first give preliminaries about MoCo, which is adopted as our baseline. Then, we introduce the design of combinatorial patches, which boost both the learning process and performance. Finally, we discuss how the proposed approach will affect the performance and computation.

\subsection{Preliminaries about MoCo}
MoCo is a highly recognized framework for self-supervised learning, which has three versions, i.e., MoCo~\cite{he2020momentum}, MoCo v2~\cite{chen2020improved}, and MoCo v3~\cite{chen2021empirical}, which gradually incorporate some of the best practice in the area. 
Specifically, MoCo v3 pipeline has two branches, i.e., an online branch and a target branch. The online branch consists of an encoder $f$ (e.g., ResNet50), a projector $g$, follow by a predictor $q$. The target branch only contains the encoder and projector with the same structure as in the online branch and its parameters are updated through an exponential moving average process as follows: 
\begin{equation}
\theta_t^f \leftarrow \alpha\theta_t^f + (1 - \alpha) \theta_o^f, \ \ 
\theta_t^g \leftarrow \alpha\theta_t^g + (1 - \alpha) \theta_o^g,
\end{equation}
where $\theta_o^f$ and $\theta_o^g$ are parameters for encoder and projector in the online branch, $\theta_t^f$ and $\theta_t^g$ are parameters for encoder and projector in the target branch. This asymmetric architecture design and the use of moving average for target branch parameters updating have been shown to help the model avoid collapse~\cite{grill2020bootstrap}. 

Given an image $x$, two different views are generated through two different augmentations $a$ and $a^\prime$, 
which are then forward to the encoders in the online and target branches respectively to retrieve the encoded embeddings as a positive pair ($v_{o}^a$, $v_{t}^{a'}$). These embeddings are then projected to vectors $z_{o}^a = q(g(v_{o}^a; \theta_{o}^g); \theta_{o}^q)$ and $z_{t}^{a'} = g(v_{t}^{a'}; \theta_{t}^g)$. Finally, the loss function for this pair $(z_{o}^{a}, z_{t}^{a'})$ is formulated by InfoNCE~\cite{van2018representation} as follows:
\begin{equation}
\mathcal{L}_{ctr}(z_{o}^{a}, \mathbf{z}_t^{a'}) = -log \frac{exp(z_{o}^{a} \cdot z_{t}^{a'}/\tau)}{\sum\limits_{z \in \mathbf{z}_t^{a'}}exp(z_{o}^{a} \cdot z/\tau)},\\
\label{eq:infonce}
\end{equation}
where $\mathbf{z}_t^{a'}$ denotes the set of target representations for all images in the batch. Note that vectors $z$, $z_o^a$, and $z_t^{a'}$  are $l_2$ normalized before computing the loss. Besides, for every sample image $x$, this loss is symmetrized as: 
\begin{equation}
\mathcal{L}_{x} =  \frac{1}{2}(\mathcal{L}_{ctr}(z_{o}^{a}, \mathbf{z}_{t}^{a^\prime}) + \mathcal{L}_{ctr}(z_{o}^{a^\prime}, \mathbf{z}_{t}^{a})).
\end{equation}

\begin{figure*}[t]
    \centering
    \includegraphics[height=3.55cm]{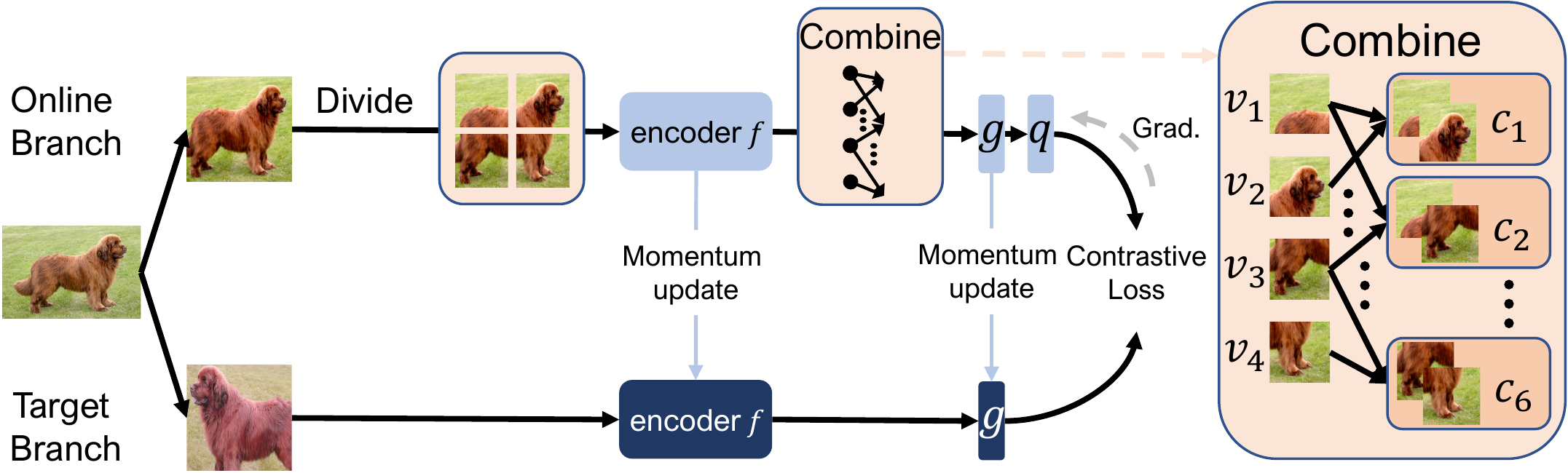}
    \caption{Overview of Fast-MoCo framework. It consists of four steps: 1) \emph{Divide} step, where the input image in the online branch is divided into multiple patches; 2) \emph{Encode} step, which the encoder $f$ encodes the features of the patches separately; 3) \emph{Combine} step, which combines the encoded features (at the last layer of the neural network); 4) the combined features are fed into projector $g$, predictor $q$, and contrastive loss for contrastive learning. Compared with MoCo, we add the Divide step and Combine Step in the online branch, with details in Section~\ref{Sec:FastMoco}. The target branch is the same as MoCo.}
    \label{fig:search}
\end{figure*}

\subsection{Fast-MoCo}\label{Sec:FastMoco}
In this section, we introduce Fast-MoCo, a simple method that can greatly improve the training efficiency of self-supervised learning with negligible extra cost. An overview of Fast-MoCo is shown in  Fig.\ref{fig:search}.
With MoCo v3 as the baseline, Fast-MoCo only makes three modifications, 1) add a \emph{Divide }step to divide an image into multiple patches before sending the patches to the encoder \footnote{In this paper, we only explore the ResNet50 as the encoder while leaving the evaluation of ViT version MoCo v3 as our future work.} of the online branch, 2) insert a \emph{Combine} step (e.g., Combine) immediately behind the encoder to combine patches, and 3) a slightly modified definition of positive and negative pairs corresponding to the divide and combine operations. In the following, we illustrate the Divide step, Combine step, and the modified loss function in detail.

\textbf{Divide Step.} For the online branch, instead of directly feed the given the augmented image $x^{a}$ into the encoder, we first divide it into  a $m \times m$ grid of patches $\{x_{p}| p \in \{1, \ldots, m^2\}\}$ as shown in Fig.\ref{fig:search}, with $\mathbf{p}$ denotes the set of patch index $\{p\}$.
The influence of $m$ in will be analyzed in Section~\ref{sec:ablation}.

\textbf{Combine Step.} Instead of directly using the encoded embedding of each patch individually for further step, we combine multiple (less than $m^2$) patch embeddings $v_p$  to form combined embeddings $c$ before sending them to further step, i.e., the projector. 

To form a combined embedding, we take a subset of $n$ indices from the patch index set $\mathbf{p}$, noted as $\mathbf{p}_{n}$($\subseteq \mathbf{p}$), and collect their corresponding features $\mathbf{v}_{\mathbf{p}_{n}} = \{v_p | p \in \mathbf{p}_{n}\}$. While there could be diverse options to combine multiple embeddings (e.g., concatenate, sum), we empirically found that simply averaging the selected features works reasonably well and is computationally efficient. Thus, in the Combine step, we generate the combined embedding by:
\begin{equation}
c = \frac{1}{n} \sum_{p \in\mathbf{p}_{n}} v_{p}. \label{eq:avgpatch}\\
\end{equation}
To improve the sample utilization efficiency, we take all possible $n$-combinations of patch embeddings for supervisions, leading to the combined embedding set $\mathbf{c} = \{c_i|i\in\{1, \ldots, C_{m^{2}}^n\}\}$, where $C_{m}^{n} = \frac{m!}{n!(m-n)!}$. 
In this way, we can generate many samples by the averaging operation in Eq.~\ref{eq:avgpatch} with negligible extra cost and ensure the sample and the target have a sufficient information gap since the combined patches embedding only covers part of the image information.

After the Combine step, the projector and the predictor in the online branch transfer each combined embedding ${c}$ to vector $z_{o}^{c}$ in a sequential manner. On the other hand, the target branch maps the other input view to ${z_{t}^{a}}'$ in the same manner as the basic MoCo v3 without modification.  
They are then L2-normalized and used for computing contrastive loss.

\textbf{Loss Functions.}
Like MoCo v3, we still utilize the contrastive loss (Eq.~\ref{eq:infonce}) to optimize the encoder, projector, and predictor. Compared with MoCo v3, Fast-MoCo does not include any extra parameters to be learned, the only difference is that there are multiple ($C_{m^2}^n$) combined patch embeddings $z_o^c$ instead of one image embedding ${z_{o}^{a}}$  corresponding to a target branch image embedding ${z_{t}^{a}}'$. We directly adapt the original loss function by averaging the contrastive losses from $C_{m^2}^n$ positive pairs between  the combined patch embeddings $z_o^c$ and target image embedding $z_{t}$. Similarly, the negative pairs are defined between the combined patch embedding and the embedding of other images in the target branch.

\subsection{Discussion}
In this section, we present some intuitive analysis about why Fast-MoCo can improve training efficiency, which will be further demonstrated with empirical results in Section~\ref{sec:experiment}.
The primary component that makes Fast-MoCo converge faster is the utilization of a set of combined patch embeddings, which significantly increase the number of positive pairs. Take $m=2$ and $n=2$ as an example, Fast-MoCo will divide the input image in the online branch into four patches and then combine their four embeddings into six, each of which represents two patches, directly expanding the number of positive pairs six times more than MoCo v3. Thus, Fast-MoCo can get more supervision signals in each iteration compared to MoCo v3 and thus achieves promising performance with fewer iterations. 

At the same time, the introduced operations in Fast-MoCo, i.e., divide an image into patches and average the representation of several patches, are extremely simple and only require negligible extra computation. The major computational cost is introduced by additional forwards over the projector and the predictor in the online branch. However, they only involve the basic linear transformations, which contributes little cost when compared to the backbone. Thus, the total overhead of Fast-MoCo accounts for $7\%$ extra training time compared to MoCo v3 (38.5 hours on 16 V100 GPUs for 100 epochs, by contrast, MoCo v3 costs 36 hours under the same setting)

Besides, since the combined patch embeddings only contain part of the information in the whole image, pulling the partially combined patches closer to the target view that contains the whole image information is more challenging than pulling the original image pairs and implicitly increasing the asymmetric of the network structure, which have been demonstrated beneficial for increasing the richness of feature representations and improve the self-supervised learning performance~\cite{grill2020bootstrap,dwibedi2021little,koohpayegani2021mean}. Owing to these merits, Fast-MoCo can achieve high sample utilization efficiency with marginal extra computational cost and thus obtain promising performance with much less training time. 
Experimental results in Section~\ref{sec:pea} and~\ref{sec:ablation} below will validate these analysis.

\section{Experimental Results}
\label{sec:experiment}

\subsection{Implementation Details}
\label{sec:details}
The backbone encoder $f$ is a ResNet-50~\cite{he2016deep} network excluding the classification layer. Following SimSiam~\cite{chen2021exploring} and MoCo v3~\cite{chen2021empirical}, projector $g$ and predictor $h$ are implemented as MLP, with the detailed configuration identical to~\cite{chen2021exploring}.
For self-supervised pretraining, we use SGD optimizer with batch size 512, momentum 0.9, and weight decay $1e^{-4}$. The learning rate has a cosine decay schedule from 0.1 to 0 with one warm-up epoch starting from 0.025. We use the same augmentation configurations as in SimSiam~\cite{chen2021exploring} (see supplementary material).

\subsection{Results}

\begin{table}[t]
	\begin{center}
	    \begin{tabular}{p{4.2cm}|>{\centering}p{1.4cm}|>{\centering}p{1.4cm}|>{\centering}p{1.4cm}|>{\centering}p{1.4cm}|>{\centering}p{1.4cm}}

		    Method 	 	& 100 ep. & 200 ep.&400 ep.&800 ep.&1000 ep.	 \tabularnewline
			\hlineB{2.5}
			SimCLR~\cite{chen2020simple} 		&64.8&67.0&68.3&69.1&-\tabularnewline
			MoCo v2~\cite{chen2020improved}       	&-&67.5&-&71.1&-\tabularnewline
			BYOL~\cite{grill2020bootstrap}      	&66.5&70.6&73.2&-&74.3\tabularnewline
			SwAV~\cite{caron2020unsupervised}     		&-&-&70.1&-&-\tabularnewline
			BarlowTwins~\cite{zbontar2021barlow}       		&-&-&-&-&73.2\tabularnewline
			SimSiam~\cite{chen2021exploring}            & 68.1&70.0 &70.8&   71.3&-\tabularnewline
			MoCo v3~\cite{chen2021empirical}      &-&-&-&73.8&-\tabularnewline
			NNCLR~\cite{dwibedi2021little}             & 69.4&70.7 &74.2&  74.9&75.4\tabularnewline
            OBoW~\cite{gidaris2021obow}        		&-&73.8&-&-&-\tabularnewline
            \textbf{Fast-MoCo} 	&\textbf{73.5}&\textbf{75.1}&\textbf{75.5}&-&-\tabularnewline
			\hline
            SwAV~\cite{caron2020unsupervised} (w/ \texttt{multi-crop})       		&72.1&73.9&-&75.3&-\tabularnewline
            DINO~\cite{caron2021emerging} (w/ \texttt{multi-crop})        	&-&-&-&75.3&-\tabularnewline
            NNCLR~\cite{dwibedi2021little} (w/ \texttt{multi-crop})            &- & -&- &  \textbf{75.6}&-\tabularnewline

		\end{tabular}
	\end{center}
	\caption{\textbf{ImageNet-1k linear evaluation results} for existing methods and our Fast-MoCo using ResNet-50. Best results are in \textbf{bold}. Fast-MoCo can achieve similar performance as MoCo v3 with only 100 epochs. When trained for 200 epochs, Fast-MoCo performances better than MoCo v3 trained for 800 epochs and is comparable with state-of-the-arts (\texttt{multi-crop} is not used in Fast-MoCo for a fair comparison).}
	\label{tab:linearIN}
\end{table}

\textbf{ImageNet Linear Evaluation}. Following \cite{chen2020simple,chen2021exploring,grill2020bootstrap}, we evaluate our method with a linear classifier on top of frozen embeddings obtained from self-supervised pretraining. The classifier is finetuned with LARS optimizer~\cite{you2017large} with configurations same as SimSiam~\cite{chen2021exploring} excepting the learning rate which we set as $lr = 0.8$.
We compare with existing methods in Table~\ref{tab:linearIN}, Our Fast-MoCo achieved 75.5$\%$ linear evaluation result with only 400 epochs of training, which shows obvious improvement of our Fast-Moco compared with all methods using two augmented views for supervision. When considering the same amount of training epoch, our result also surpass SwAV~\cite{caron2020unsupervised} and DINO~\cite{caron2021emerging} even including the use of \texttt{multi-crop}~\cite{caron2020unsupervised}. Note that our new design is orthogonal to \texttt{multi-crop}~\cite{caron2020unsupervised} (details in Section~\ref{sec:mc}) and the novel designs in SwAV, DINO and NNCLR.

\textbf{Semi-Supervised Learning}.
\begin{table}[t]
	\begin{center}
	    \begin{tabular}{p{3.8cm}|>{\centering}p{1cm}|>{\centering}p{1cm}|>{\centering}p{1cm}|>{\centering}p{1cm}}
		    \multirow{2}{4em}{Method} 	 	& \multicolumn{2}{c|}{1\%} &\multicolumn{2}{c}{10\%}	 	\tabularnewline
		             	 	& Top-1 & Top-5&Top-1&Top-5	 	\tabularnewline
			\hlineB{2.5}

			Supervised 		&25.4&48.4&56.4&80.4\tabularnewline
			\hline
            InstDisc~\cite{wu2018unsupervised}       		&-&39.2&-&77.4\tabularnewline
            PIRL~\cite{misra2020self}        	&-&57.2&-&83.8\tabularnewline
            SimCLR~\cite{chen2020simple}        	&48.3&75.5&65.6&87.8\tabularnewline
            BYOL~\cite{grill2020bootstrap}        	&53.2&78.4&68.8&89.0\tabularnewline
      
            Barlow Twins~\cite{zbontar2021barlow}        	&55.0&79.2&69.7&89.3\tabularnewline
            NNCLR~\cite{dwibedi2021little}        	&56.4&80.7&69.8&89.3\tabularnewline
            \textbf{Fast-MoCo}        	&\textbf{56.5}&\textbf{81.1}&\textbf{70.3}&\textbf{89.4}\tabularnewline
			\hline
            SwAV~\cite{caron2020unsupervised} (w/ \texttt{multi-crop})        	&53.9&78.5&70.2&89.9\tabularnewline
		\end{tabular}
	\end{center}
	\caption{\textbf{Semi-supervised learning results on ImageNet-1K} with ResNet-50 backbone. We report Top-1 and Top-5 accuracies for models finetuned with $1\%$ and $10\%$ labeled data. 
	Detailed configuration can be found in supplementary material.}
	\label{tab:semi}
\end{table}
Following the semi-supervised learning setting in~\cite{chen2020simple}, we fine-tune our model pretrained by 400 epochs with $1\%$ and $10\%$ of the data split. The results are shown in Table~\ref{tab:semi}.  Our  method  performs better than all compared methods w/o \texttt{multi-crop} and is on par with SwAV using \texttt{multi-crop}. 

\textbf{Transfer Learning}.
Table~\ref{tab:det} shows experimental results evaluating the effectiveness of the learned model when transferred to detection and segmentation tasks. For object detection on PASCAL-VOC~\cite{everingham2010pascal}, with Faster R-CNN~\cite{ren2015faster} framework, we have all weights finetuned on the \texttt{trainval07+12} dataset and evaluated on the \texttt{test07} dataset. For detection and instance segmentation on COCO~\cite{lin2014microsoft}, we finetune our weights with Mask R-CNN~\cite{he2017mask} on the \texttt{train} set and report results on the \texttt{val} split. The results in Table~\ref{tab:det} show that our Fast-MoCo performs on par with or better than the state-of-the-arts in localization tasks.

\section{Analysis}
\subsection{Same or Different Augmented Views} 
Recent works~\cite{chen2020simple,grill2020bootstrap} have indicated that contrastive methods are sensitive to augmentations, especially spatial transformations~\cite{chen2020simple}. Compared with the conventional settings of having different augmented view (73.5\% on ImageNet for 100-epoch training of Fast-MoCo), we observe severe drop of accuracy (48.5\%) if the positive embedding pair in Eq.~(\ref{eq:loss_full}) are from the same augmented view, i.e. $a' = a$.
When the same augmented view is used, the detrimental non-semantic information contained in patches would be exposed to its contrastive target, which causes the significant drop of accuracy. 
These results show the importance of using appropriate targets for contrastive learning.

\begin{table*}[t]
	\begin{center}
	    \begin{tabular}{p{3.8cm}|>{\centering}p{0.8cm}|>{\centering}p{0.8cm}|>{\centering}p{0.8cm}|>{\centering}p{0.8cm}|>{\centering}p{0.8cm}|>{\centering}p{0.8cm}|>{\centering}p{0.8cm}|>{\centering}p{0.8cm}|>{\centering}p{0.8cm}}
		    \multirow{2}{4em}{Method}  	 	& \multicolumn{3}{c|}{VOC det} & \multicolumn{3}{c|}{COCO det}	&\multicolumn{3}{c}{COCO seg}	\tabularnewline
		             	 	& $AP_{all}$ & $AP_{50}$&$AP_{75}$&$AP_{all}^{bb}$&$AP_{50}^{bb}$&$AP_{75}^{bb}$
		             	 	&$AP_{all}^{mk}$&$AP_{50}^{mk}$&$AP_{75}^{mk}$\tabularnewline
			\hlineB{2.5}

			Supervised 		&53.5&81.3&58.8&38.2&58.2&41.2&33.3&54.7&35.2\tabularnewline
			\hline
            MoCo V2~\cite{chen2020improved}       		&57.4&82.5&64.0&39.3&58.9&42.5&34.4&55.8&36.5\tabularnewline
            SimSiam~\cite{chen2021exploring}        	&57&82.4&63.7&39.2&\textbf{59.3}&42.1&34.4&\textbf{56.0}&36.7\tabularnewline
            Barlow Twins~\cite{zbontar2021barlow}        	&56.8&82.6&63.4&39.2&59.0&42.5&34.3&\textbf{56.0}&36.5\tabularnewline
            \textbf{Fast-MoCo}        	&\textbf{57.7}&\textbf{82.7}&\textbf{64.4}&
                                \textbf{39.5}&59.2&\textbf{42.6}&
                                \textbf{34.6}&55.9&\textbf{36.9}\tabularnewline
            \hline                                
            SwAV~\cite{caron2020unsupervised} (w/ \texttt{multi-crop})  &56.1&82.6&62.7&38.4&58.6&41.3&33.8&55.2&35.9\tabularnewline                                
		\end{tabular}
	\end{center}
	\caption{\textbf{VOC and COCO object detection (det) and instance segmentation (seg) results}. We report results measured by Average Precision (AP) using ResNet50 with the C4 backbone variant~\cite{girshick2018detectron}. For VOC dataset, we train on \texttt{trainval07+12} and evaluate on \texttt{test07} by running three trials and report the averaged results.}
	\label{tab:det}
\end{table*}

\subsection{Comparison on Patch Encoding Approaches}
\label{sec:pea}
Apart from our proposed Fast-MoCo pipeline, there is also a number of alternatives~\cite{noroozi2016unsupervised,van2018representation,henaff2020data,gidaris2021obow,chen2021jigsaw,pathak2016context,bao2021beit,he2021masked} that falls into the same category with our Fast-MoCo which does not apply the \emph{two-image-one-pair} mechanism. In this Section, we provide a detailed comparison on these variants.

\textbf{Sample-Encode-Combine.}
The compared settings contain cases where patches can not be generated from dividing a $224\times224$ view. Apart from the Fast-MoCo baseline, we set up a Sample-Encode-Combine (SEC) configuration for fair comparison. In SEC configuration, we replace the 'Divide' step in Fast-MoCo by randomly and independently sampling patches. In contrast to Fast-MoCo with $2\times4$ patches divided from two $224\times224$ views, for SEC we have eight independently sampled patches :$\{x_{p}| p \in \{1, \ldots, 8\}\}$ and two $224\times224$ target $\{x_{t}^{a}, x_{t}^{a'}\}$. 
As $x_{p}$ for SEC are not devided from the target views $x_{t}$. The embeddings of all eight $x_{p}$ can be combined with each other to get combined embedding $c$, we have the amount of combination increased from $2C_{4}^{2}=12$ to $C^{2}_8=28$. The loss function for SEC is written as follows:
\begin{equation}
\mathcal{L}_{x} = \frac{1}{2C^{2}_8}\sum\limits_{c\in\mathbf{c}}(\mathcal{L}_{ctr}(z_{c}, \mathbf{z}_{t}^{a}) + \mathcal{L}_{ctr}(z_{c}, \mathbf{z}_{t}^{a'})), \\
\label{eq:loss_full}
\end{equation}
It obtains $72.8\%$, which is the second-best among all variants in Table~\ref{tab:dual}(a).

\begin{table*}[t]
    \begin{subtable}[h]{0.5\textwidth}
	    \centering
	    \footnotesize
		\begin{tabular}{l|c|c}
		    \multirow{2}{4em}{Method}  & Num. of 	&  \multirow{2}{3em}{Top-1}	 	\\
		    &Samples &\\
		    \hlineB{2.5}
			Encode Only	&4	&68.9\\
			\hline
			Sample-Combine-Encode	&4	&71.2\\
			Divide-Combine-Encode	&4	&71.8\\
		    \hline
			Montage-Encode-Divide-	&\multirow{2}{1em}{28}	&\multirow{2}{1.8em}{70.4}\\	
			Combine	&&\\	
			\hline
			Sample-Encode-Combine	&28	&72.9\\	
			\textbf{Fast-MoCo}	&12	&73.5\\				
		\end{tabular}
	    \caption{\textbf{Comparison of patch encoding approaches.} Results are based on ImageNet linear evaluation, all models are pretrained for 100 epochs.}
    \end{subtable}
    \hfill
    \begin{subtable}[h]{0.45\textwidth}
    \centering
    	\begin{tabular}{l|c|c|c}

		    Case 	 	&  \texttt{multi-crop} & Comb. & Top-1	 	\\
		    \hlineB{2.5}
			MoCo v3		&-&-&70.3\\
			(i)		&\cmark&-&73.1\\
			(ii)		&-&\cmark&73.5\\
			(iii)		&\cmark&\cmark&74.2\\			

		\end{tabular}
        \caption{\textbf{Relationship with \texttt{multi-crop}.} `Comb.' denotes the usage of combinatorial patches. Results are linear evaluation on ImageNet, all models are pretrained for 100 epochs.}		
    \end{subtable}
\caption{}
\label{tab:dual}
\end{table*}

\textbf{Encode Only.} A widely adopted way to encode patches is to encode them separately~\cite{noroozi2016unsupervised,van2018representation,henaff2020data,gidaris2021obow}, which do not include the `Divide' step or `Combine'  step in our Fast-MoCo as depicted in Fig.~\ref{fig:search}. For a fair comparison, the patch used for encoding should contain approximately the same amount of information as two $112\times112$ patches combined, so we set the spatial size of the patch as $158\times158$. In doing so, we can no longer retrieve these patches by dividing a $224\times224$ that we use for contrastive target, thus they are independently generated by augmentation as described in Section \ref{sec:details}. We generate four $158\times158$ patches $\{x_p\}$ and two $224\times224$ target $\{x_{t}^{a}, x_{t}^{a'}\}$, 
for each image $x$ we have:
\begin{equation}
\mathcal{L}_{x} = \frac{1}{8}\sum\limits_{z_p\in\mathbf{z}_p}(\mathcal{L}_{ctr}(z_{p}, \mathbf{z}_{t}^{a}) + \mathcal{L}_{ctr}(z_{p}, \mathbf{z}_{t}^{a'})),\\
\label{eq:se}
\end{equation}
where $\mathbf{z}_{target}$ denotes the target vectors in a mini-batch and $\mathbf{z}_p$ denotes the features of the four patches sampled from the image $x$. As shown in Table~\ref{tab:dual}(a), the result of Encode Only is $68.9\%$.

\textbf{Divide(Sample)-Combine-Encode.} While Fast-MoCo encodes the small divided patches independently and combines them at embedding level; one can also combine them at image level with patches placed in their original positions, thus preserving the relative positional information among patches.
Note that if the stitched image is not in a rectangular shape, the redundant computational cost would be hard to avoid for a CNN encoder. In the Divide step, we divide a $224 \times 224$ image vertically and horizontally to get four $112\times112$ patches. In the Combine step for Divide-Combine-Encode, two $112\times112$ patches are stitched to $112\times224$ or $224\times112$ at image level. The Divide step, Encode step, and losses are the same as Fast-MoCo. As shown by Divide-Combine-Encode in Table~\ref{tab:dual}(a), compared to Encode Only with four squared $158\times158$ crops, these rectangular crops with less locally-bounded features is preferred with a +2.9 gain. 
Divide-Combine-Encode can also be viewed as bringing the Combine step of our Fast-MoCo pipeline before the encoding step. 
Compared with the Fast-MoCo pipeline, 1) the Fast-MoCo Divide-Combine-Encode pipeline generates fewer target-sample pairs for the same computational cost, and 2) does not include sufficiently difficult target-sample pairs (more discussion in Section ~\ref{sec:ablation}).

For the \emph{Sample-Combine-Encode} in Table~\ref{tab:dual}(a), we generate the $112\times112$ rectangular patches independently, and find its +2.3 gain over Encode Only. Sample-Combine-Encode performs worse than Divide-Combine-Encode because the divided patches in Divide-Combine-Encode have no overlap, which maximizes the diversity of the combined patches, but Sample-Combine-Encode cannot guarantee non-overlapping patches.

\textbf{Montage-Encode-Divide-Combine.}
JigClu~\cite{chen2021jigsaw} proposed a patch encoding technique with montage image. Given a batch of $K$ images, four patches are generated from each image with different augmentations, resulting in a mini-batch of $4K$ patches. Then $K$ montage images of size $224\times224 $ are generated by stitching four patches randomly selected (without replacement) from the mini-batch of $4K$ patches. The encoder adds an additional step before average pooling to divide $K$ montage feature maps back to $4K$ patch features to get their encoded embeddings. We replaced our Divide-Encode steps with this Montage-Encode-Divide approach, forming a Montage-Encode-Divide-Combine pipeline. The result of this approach in Table~\ref{tab:dual}(a) shows that it is not as good as the relatively simpler Fast-MoCo approach.

\textbf{Analysis}
All in all, our Fast-MoCo outperform other variants with a steady margin. The Encode Only baseline achieves $68.9\%$. If we combine inputs before the encoding mechanism, the performance improved to $71.2\%$ and $71.8\%$ for inputs obtained by random cropping and dividing respectively. If we combine the embedding after encoding inputs, the performance improved to $72.9\%$ (sample by random cropping) and $73.5\%$ (Fast-MoCo). The Montage strategy achieves $70.4\%$. We find that the Sample (random cropping) always performs worse than Divide, and combine after encoding always better than before encoding in our experiments. Based on these results, we found non-overlapping patches(Divide) and Combine after encoding to be the best practice.

\subsection{Relationship with Multi-Crop} 
\label{sec:mc}
\texttt{Multi-crop} is a technique proposed in SwAV~\cite{caron2020unsupervised}. In addition to two $224\times224$ crops, \texttt{multi-crop} additionally adds six $96\times96$ patches as samples so that the encoder is trained with samples that have multiple resolutions and hard samples. However, the additional samples also needs more computation. 
While both Fast-MoCo and \texttt{mulit-crop} use small patches as their input, Fast-MoCo is not trained with samples of multiple resolutions. Except the (iii) in Table~\ref{tab:dual}(b), all reported results in this paper for Fast-MoCo are w/o \texttt{mulit-crop}. 
Nevertheless, as shown by (ii) in Table~\ref{tab:dual}(b),  Fast-MoCo w/o \texttt{mulit-crop} is 0.4 better than MoCo v3 w/ \texttt{mulit-crop}. 
Fast-MoCo w/ \texttt{mulit-crop} (see supplementary material for details), i.e. (iii) in Table~\ref{tab:dual}(b), further improves the result of Fast-MoCo by 0.7, which shows that our contribution is orthogonal to \texttt{mulit-crop}.

\begin{figure*}[t]
    \centering
    \begin{subfigure}[b]{0.27\textwidth}
        \centering
        \includegraphics[width=\textwidth]{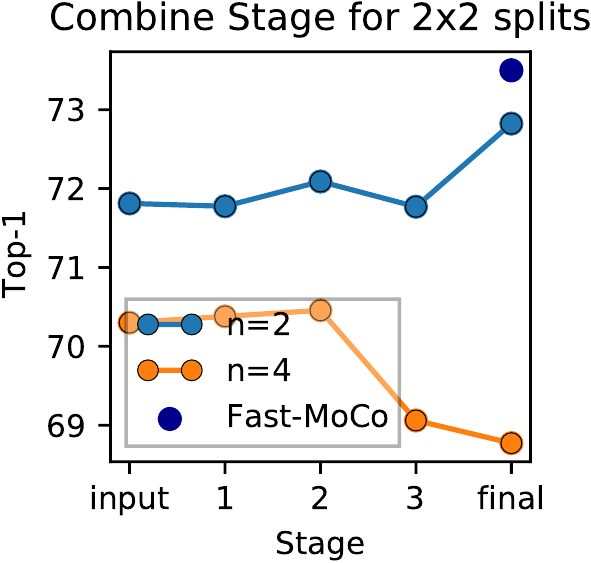}
        \caption{}
    \end{subfigure}
    \hfill
    \begin{subfigure}[b]{0.55\textwidth}
        \centering
        \includegraphics[width=\textwidth]{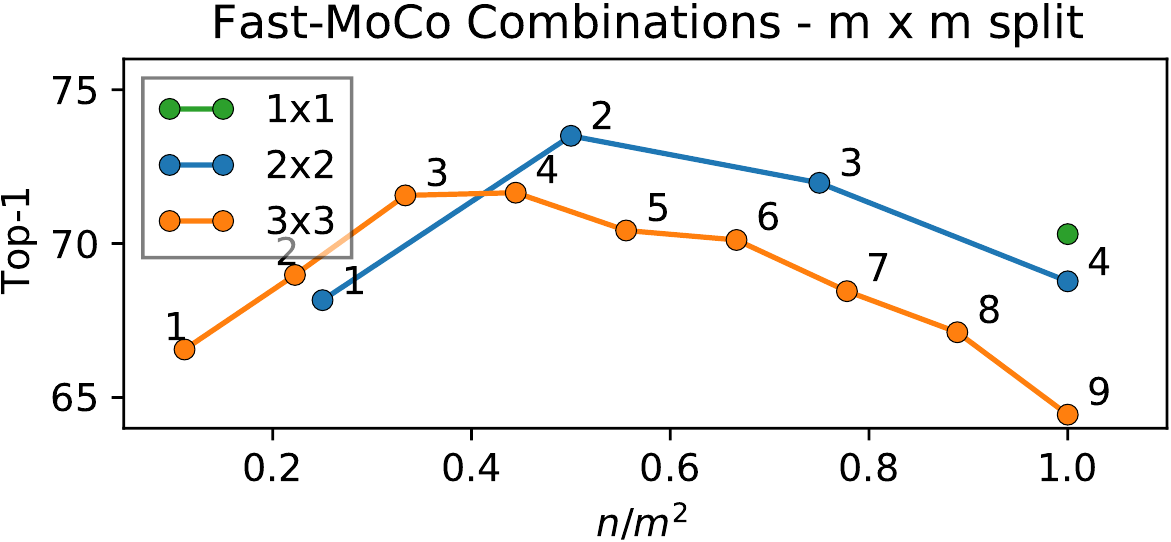}
        
        \caption{}
    \end{subfigure}
    \caption{(a): ImageNet linear evaluation accuracy (Y-axis) when different ResNet stages (X-axis) are selected for combining $n=2$ divided patches or $n=4$ divided patches in the Divide step. (b): ImageNet accuracy (Y-axis) when $n/m^2$ (X-axis) patches are combined for $m\times m$ ($1\times1$, $2\times 2$, $3\times 3$) divided patches. Annotations represent the number of combined samples $n$.}
    \label{fig:ablate}
\end{figure*}

\subsection{Ablation on Fast-MoCo}
\label{sec:ablation}
\textbf{Combine Stage and Task Difficulty}
In our Fast-MoCo pipeline, a $224\times 224$ cropped image is divided into four patches. The embeddings of these four patches  are combined at the final layer of the ResNet encoder. In this Section, we investigate the influence of combining $n=2$ patches or $n=4$ patches. When $n=2$, there is an information gap between sample and target because the sample only has half of its patches used for contrastive loss. When $n=4$, all information within the original image is preserved.
When combining two patches (or their feature maps) before the last stage, as it is difficult to handle non-rectangle input for CNN, we only stitch them vertically and horizontally with respect to their original position as described in Section~\ref{sec:pea}. Since convolution layers are computationally heavy, we do not reuse patches/patch feature maps, so uniformly, we have two target-sample pairs per image when $n=2$. In the case of the final layer, for a fair comparison, we adopt the same sample pair selected as in previous stages, which means two target-sample pairs per image.

In Figure~\ref{fig:ablate}(a), the results show that
when the Combine step took place at the embedding level, i.e., the elimination of relative positional information between patches at later stages, it is beneficial when there is an information gap between sample and target ($n=2$). However, it will be harmful when there is no gap ($n=4$). 
On the other hand, we can see the training does benefit from a harder task, i.e., presence of information gap between sample and target when $n=2$.
While for our Fast-MoCo, it will further improve the result as more samples are generated with the help of embedding level combination. 

\textbf{Number of Combined Samples} 
Given $m^2=4$ divided patches and $n=2$ patches to be combined, we have $C_{m^{2}}^n=6$ target-sample pairs, but is it necessary to use them all? From these 6 target-sample pairs, when we use 2, 4, and 6 target-sample pairs per image and ensure all patches are selected for combination for equal times, the accuracy are 72.6,  73.3, and 73.5, respectively. 
These results show that more samples from combination helps to learn better representations.

\textbf{Number of Divided and Combined Patches}
Figure~\ref{fig:ablate}(b) shows the influence of choosing different numbers of divided patches $m\times m$ and numbers of combined patches $n$. 
The performance is controlled by two factors: 1) the divide base number $m$, which determines the patch size, and 2) the percentage of the covered area by selected patches combined, i.e., $n/m^2$. 
With a proper selection of $n/m^2$ by controlling $n$, we can benefit from extra samples and difficulty it self. Meanwhile, making the task too hard with $n/m^2$ close to 0  (e.g. $n=1$ for $m^2=2\times 2$), or making the actual patches too small, e.g. $3\times 3$ are both harmful to the performance. We find choosing 2x2 split with $n=2$ have a good trade-off for these factors, which is used for our key results. When $n$ is close to the optimal choice, i.e. $n=2$ for $m^2=2\times 2$ or $n=3$ for $m^2=3\times 3$, the small variation of $n$ (e.g. $n=4$ for $m^2=3\times 3$) does not lead to large variation of ImageNet top-1 accuracy, showing Fast-MoCo is relatively stable to the variation of $n$ and $m$.

\subsection{Combination Method}
In this section we discuss different combination choices in the Combine step. We consider two alternatives: weighted average and merge by max operation. 

\subsubsection{Weighted Average.}
Consider the case of combining 2 patches $p$ and $p'$ from the $2\times 2$ divided patches, for patch embeddings $v_{p}$ and $v_{p'}$ of patches $p$ and $p'$ respectively, we have:
\begin{equation}
c = \gamma v_{p} + (1 - \gamma) v_{p'}, \\
\end{equation}
where $p' \neq p$ and every patch is selected for equal times. By adjusting $\gamma$ within the range of $[0.5, 1)$, we create a continuous transition between using patch embeddings separately and combinatorial patches with four combinations. The results are shown in Figure~\ref{fig:combination}(a), from which we can see the best setting is to have $\gamma=0.5$, which assigns equal weights for both patches. Therefore, equal weight for Fast-MoCo is the default setting in other experimental results.
The transition is idiosyncratic when the weight for either feature is close to zero. 

\subsubsection{Weighted Average with Weight from Random Sampling.}
\begin{figure*}[t]
    \centering
    \begin{subfigure}[b]{0.48\textwidth}
        \centering
        \includegraphics[width=\textwidth]{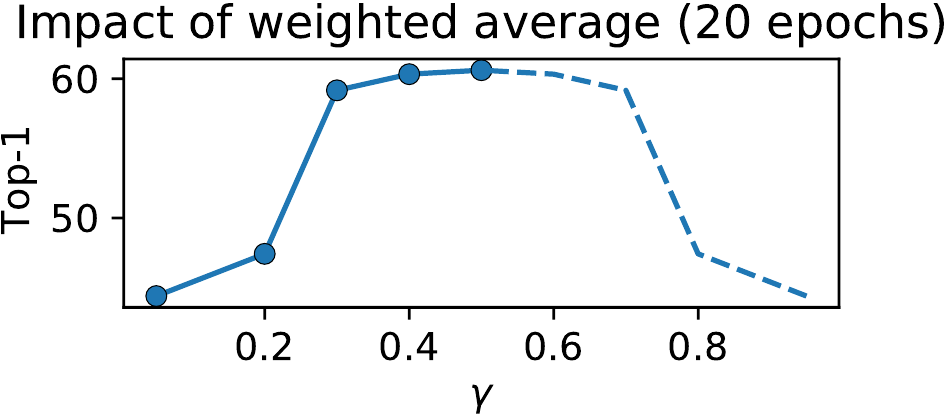}
        \label{CRFsearchspace}
        \caption{}
    \end{subfigure}
    \begin{subfigure}[b]{0.48\textwidth}
        \centering
        \includegraphics[width=\textwidth]{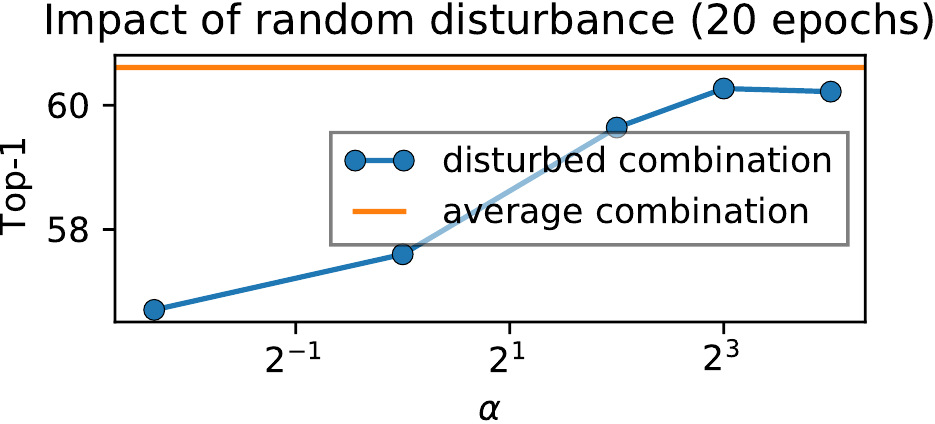}
        \label{prunespace}
        \caption{}
    \end{subfigure}
    \caption{(a): Random weighted average - fix value. (b): Random weighted average - Beta distribution.}
    \label{fig:combination}
\end{figure*}

Apart from weighted combining with fixed weights, we also investigated the case when $\gamma$ is randomly sampled from beta distribution; we have $\gamma \sim Beta(\alpha, \alpha)$ with $\alpha \in \{0.2, 1, 4, 8, 16\}$. As shown in Figure~\ref{fig:combination}(b), The result gradually approaches average combination as randomness is suppressed by higher $\alpha$. We conclude that the combination of patch embedding is best done with its patch members contributing equally to the combined embedding.

\subsubsection{Max Operation.}
As for combination with max operation, for each feature channel $i$, we have:
\begin{equation}
c^{(i)} = \max\limits_{\substack{v\in\{v_p, v_{p'}\}}} v^{(i)}. \\
\end{equation}
The 100-epoch linear evaluation result when the max operation is used at the Combine step is 64.6, which is significantly lower than the result of 73.5 for the Fast-MoCo counterpart with weighted average.

\section{Conclusion}
In this work, a simple yet effective self-supervised learning method, i.e., Fast-MoCo, is proposed to boost the training speed of the momentum-based contrastive learning method. By extending the MoCo v3 baseline with our proposed divide and combine steps, Fast-MoCo can construct multiple positive pairs with moderately more challenging optimization objectives for each input, which could significantly increase the sample utilization efficiency with negligible computational cost. Linear evaluation results on ImageNet show that Fast-MoCo trained with 100 epochs can achieve on-par performance with MoCo v3 trained with 800 epochs, which significantly lowers the computation requirements for self-supervised learning research and breaks the barrier for the general academic community. More extensive experiments and analyses further demonstrate the transferability  of Fast-MoCo to other tasks and validate our design.


\noindent\textbf{Acknowledgement.} This work was supported by the Australian Research Council Grant DP200103223, Australian Medical Research Future Fund MRFAI000085, CRC-P Smart Material Recovery Facility (SMRF) – Curby Soft Plastics, and CRC-P ARIA - Bionic Visual-Spatial Prosthesis for the Blind.

%
%
\bibliographystyle{splncs04}
\bibliography{egbib}

\clearpage
\appendix

\section{Algorithm}
\begin{algorithm}
\caption{Pytorch-style Pseudocode for Fast-MoCo}\label{alg:cap}
\begin{minted}[fontsize=\footnotesize]{python}
# f_o: online branch networks [encoder, projector, predictor]
# f_t: target branch networks [encoder, projector]
# a: exponential moving average momentum \alpha, t: temperature \tau
# combine: generate all possible 2-combinations between patch embeddings

for x in loader: # load a minibatch
    x1, x2 = aug(x), aug(x) # augemtation, NxCxHxW

    x1_d, x2_d = divide(x1), divide(x2) # Divide step, 4NxCx(H/2)x(W/2)
    v1, v2 = f_o[0](x1_d), f_o[0](x2_d) # online branch encode
    c1, c2 = combine(v1), combine(v2) # Combine step
    
    z1_c, z2_c = f_o[1:](c1), f_o[1:](c2) # project & predict
    z1, z2 = f_t(x1), f_t(x2) # target branch encode & project
    
    loss = (ctr(z1_c, z2) + ctr(z2_c, z1)) / 2
    loss.backward() 
    
    # weight update
    update(f_o.params) 
    f_t.params = a * f_t.params + (1-a) * f_o[:2].params 

def ctr(z_c, z)
    z_c, z = normalize(z_c, dim=1), normalize(z, dim=1) # l2-normalize
    z_c = z_c.split(z.size(0))
    
    # calculate loss for each of the 6 combined samples
    loss = 0
    for _z in z_c: 
        logits = mm(_z, z.t())
        loss += CorssEntropyLoss(logits/t, labels) 
        # positive pairs are sourced from the same instance
    return loss /=  len(z_c)
\end{minted}
\end{algorithm}

\section{Additional Implementation Details}
\subsection{Self-Supervised Pretraining}
Here in Table~\ref{tab:ssp}, we show the detailed configurations for self-supervised pretraining. For ablation study experiments that are only trained for 20 epochs, we multiply the learning rate by 1.5 compared to default setting.
\begin{table}[H]
    \centering
    \footnotesize
	\begin{tabular}{p{6.5cm}|>{\centering}p{2cm}>{\centering}p{2cm}}
	    \textbf{Config}  & \multicolumn{2}{c}{\textbf{Values}} 	 	\tabularnewline

	    \hlineB{2.5}
	    Epochs	&  $\{100, 200, 400\}$& $20$ 	 	\tabularnewline
	    \hline
		Optimizer	& \multicolumn{2}{c}{SGD}	\tabularnewline
		Optimizer momentum	& \multicolumn{2}{c}{0.9}	\tabularnewline
		Weight decay	& \multicolumn{2}{c}{1e-4}	\tabularnewline	
		Gradient clipping	&\multicolumn{2}{c}{1.0}\tabularnewline			
		\hline
		Learning rate schedule	& \multicolumn{2}{c}{Cosine}\tabularnewline		
		Initial Learning rate & 0.1& 0.15\tabularnewline
		Final Learning rate & \multicolumn{2}{c}{0.0}\tabularnewline

		Warmup epochs & \multicolumn{2}{c}{1}\tabularnewline
		Warmup initial learning rate & 0.025& 0.0375\tabularnewline
		\hline
		Batch size	&\multicolumn{2}{c}{512}	\tabularnewline
		Temperature $\tau$	&\multicolumn{2}{c}{1.0}\tabularnewline	
		Exponential moving average momentum $\alpha$ 	&\multicolumn{2}{c}{0.99}\tabularnewline	
		Augmentation &\multicolumn{2}{c}{As in~\cite{chen2021exploring}}\tabularnewline			
	\end{tabular}
    \caption{\textbf{Self-supervised pretraining setup.}}
    \label{tab:ssp}
\end{table}    
\subsection{Linear Evaluation}
Here in Table~\ref{tab:le}, we show the detailed configurations for linear evaluation. For ablation study experiments that only trained for 20 epochs, we have a shorter training schedule with doubled learning rate.
\begin{table}[H]
    \centering
    \footnotesize
	\begin{tabular}{p{6.5cm}|>{\centering}p{2cm}>{\centering}p{2cm}}
	    \textbf{Config}  & \multicolumn{2}{c}{\textbf{Values}} 	 	\tabularnewline

	    \hlineB{2.5}
	    Pre-trainig epochs	&  $\{100, 200, 400\}$& $20$ 	 	\tabularnewline
	    Fine-tuning epochs	&  90& 10 	 	\tabularnewline
	    \hline
		Optimizer	& \multicolumn{2}{c}{LARS}	\tabularnewline
		Optimizer momentum	& \multicolumn{2}{c}{0.9}	\tabularnewline
		\hline
		Learning rate schedule	& \multicolumn{2}{c}{Cosine}\tabularnewline		
		Initial Learning rate & 0.8& 1.6\tabularnewline
		Final Learning rate & \multicolumn{2}{c}{0.0}\tabularnewline
		\hline
		Batch size	&\multicolumn{2}{c}{4096}	\tabularnewline
		Augmentation &\multicolumn{2}{c}{As in~\cite{chen2021empirical}}\tabularnewline
	
	\end{tabular}
    \caption{\textbf{Linear evaluation setup.}}
    \label{tab:le}
\end{table} 
\subsection{Semi-Supervised Training}
We follow the evaluation protocol as in~\cite{chen2020simple,grill2020bootstrap,zbontar2021barlow} and apply the same augmentations as used in the linear evaluation. For both $1\%$ and $10\%$ settings, we adopt the same dataset split as in~\cite{chen2020simple}. We use a SGD optimizer with Nesterov momentum of 0.9 and a batch size of 256. We do not apply any regularization such as weight decay and gradient clipping. 
The learning rate is scaled by a factor of 0.2 at $60\%$ and $80\%$ of the training schedule. For the $1\%$ setting, we fine-tune for 60 epochs with a learning rate of 5e2 and 0 for the linear layer and backbone, respectively. For the $10\%$ setting, we fine-tune for 20 epochs with a learning rate of 5e2 and 1e-6 for the linear layer and backbone, respectively.

\subsection{Transfer Learning}
For object detection and segmentation, we follow the evaluation protocol as in~\cite{he2020momentum,chen2020improved,chen2021exploring,zbontar2021barlow} and conduct experiments on detectron2~\cite{wu2019detectron2} codebase with the R50-C4 backbone variant~\cite{girshick2018detectron}. The detailed configurations are as follows:
\subsubsection{PASCAL-VOC}
For object detection on PASCAL-VOC~\cite{everingham2010pascal} with Faster R-CNN~\cite{ren2015faster}, we have all weights finetuned on the \texttt{trainval07+12} dataset and evaluated on the \texttt{test07} dataset. We fine-tune for $24K$ iterations with batchsize 16. The learning rate is 1e-2 and 1e-3 for the heads and backbone, respectively, and is scaled by 0.1 at $18K$ and $22K$ iterations.
\subsubsection{COCO}
For detection and instance segmentation on COCO~\cite{lin2014microsoft}, we finetune our weights with Mask R-CNN~\cite{he2017mask} on the \texttt{train} set and report results on the \texttt{val} split. We use the $1\times$ schedule in detectron2~\cite{wu2019detectron2}. The learning rate is 1e-2 and 1e-3 for the heads and backbone, respectively.

\subsection{Fast-MoCo w/ \texttt{mulit-crop}}
\texttt{Multi-crop} is a technique proposed in SwAV~\cite{caron2020unsupervised}. In addition to two $224\times224$ crops, \texttt{multi-crop} additionally adds six $96\times96$ patches as samples so that the encoder is trained with samples that have multiple resolutions and hard samples. For fair comparison, we add an additional $224\times224$ crop as sample (which adds approximately the same computational cost as six $96\times96$ patch samples) for Fast-MoCo w/ \texttt{mulit-crop}.


\section{Additional Results}
\subsection{Combinatorial Patches on Other SSL Frameworks.}
We further evaluate our method by directly applying it to NNCLR (neighborhood-replaced target embedding), BYOL (non-contrastive), and SimSiam (momentum-free); results in Table~\ref{tab:R2} show that the proposed combinatorial patches can generalize well to different SSL frameworks. Please note, while the proposed combinatorial patches can be applied to MoCoV3, NNCLR, and BYOL seamlessly, integrating it to SimSiam will increase the computation cost since we need to forward the same view twice to get the symmetrized loss while SimSiam only needs one forward pass. Nevertheless, combinatorial patches can still boost SimSiam significantly, e.g., achieving 71.7\% top-1 linear evaluation accuracy with only 100 epochs, which is higher than the original SimSiam running 800 epochs (71.3\%).

\begin{table}[ht]
\footnotesize
	\begin{center}
		\begin{tabular}{p{8.cm}|>{\centering}p{2.2cm}}
			Method		&  Top-1 (e100)	\tabularnewline
			\hlineB{2.5}
			NNCLR~\cite{dwibedi2021little} 			&69.4\tabularnewline
			NNCLR + combinatorial patches  		 &\textbf{72.5}  (+3.1)\tabularnewline
			\hline	
			BYOL~\cite{grill2020bootstrap}  			&66.5\tabularnewline
			BYOL + combinatorial patches  			&\textbf{73.6}  (+7.1)\tabularnewline
			\hline			
			SimSiam~\cite{chen2021exploring} 			&68.1\tabularnewline
			SimSiam + combinatorial patches   				&\textbf{71.7}  (+3.6)\tabularnewline
		\end{tabular}
	\end{center}
	\caption{ImageNet linear evaluation performance of our method on other frameworks with ResNet-50. Results of compared methods w/o combinatorial patches are from their original paper.}
	\label{tab:R2}
\end{table}

\subsection{Downstream Results with Different Pretrain Epochs}
\begin{figure}[t]
    \centering
    \includegraphics[width=10.0cm]{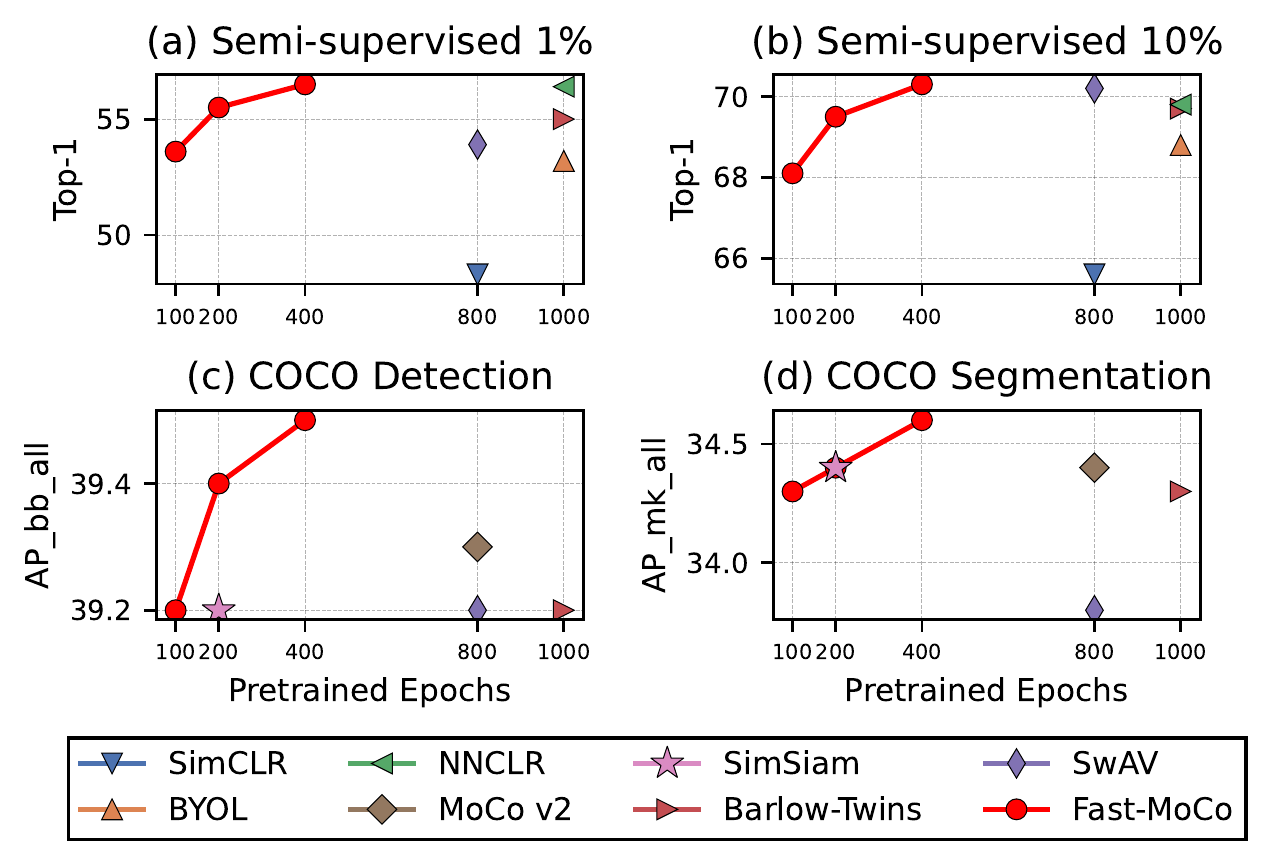}
    \caption{Downstream task results in different epochs.}
    \label{fig:Downstream}
\end{figure}

The faster convergence of our method also stand for the low-data regime and transfer learning as shown in Figure~\ref{fig:Downstream}. Fast-MoCo achieves better or similar performance with only 100 or 200 epochs than other methods trained with 800 or 1000 epochs on semi-supervised classification with 1\% labeled data (Figure~\ref{fig:Downstream}(a)) and 10\% labeled data (Figure~\ref{fig:Downstream}(b)), COCO detection (Figure~\ref{fig:Downstream}(c)), and COCO segmentation (Figure~\ref{fig:Downstream}(d)). 

\subsection{Robustness to the Selection of Batch Size}
In Table~\ref{R3}, we show the comparison of robustness to batch size varying among \{1024, 512, 256\} with the reported values in SimSiam, BYOL and Barlow Twins. As observed, Fast-MoCo has similar robustness as BYOL and Barlow Twins.

\begin{table}[ht]
	\begin{center}
        \begin{tabular}{p{3.9cm}|>{\centering}p{1cm}>{\centering}p{1cm}>{\centering}p{1cm}}
\multicolumn{1}{c}{\multirow{2}{*}{Method}} & \multicolumn{3}{c}{Top-1}                                    \tabularnewline \cline{2-4} 
\multicolumn{1}{c}{}                        & \multicolumn{1}{>{\centering}p{1cm}}{b1024} & \multicolumn{1}{>{\centering}p{1cm}}{b512} &\multicolumn{1}{>{\centering}p{1cm}}{b256}  \tabularnewline \hlineB{2.5} 
SimSiam~\cite{chen2021exploring} (100 ep.)                             & \multicolumn{1}{c|}{68.0}  & \multicolumn{1}{c|}{68.1} & 68.1 \tabularnewline 
BYOL~\cite{grill2020bootstrap} (300 ep.)                                & \multicolumn{1}{c|}{72.3}  & \multicolumn{1}{c|}{72.2} & 71.9 \tabularnewline 
Barlow Twins~\cite{zbontar2021barlow} (300 ep.)                        & \multicolumn{1}{c|}{71.7}  & \multicolumn{1}{c|}{71.4} & 70.7 \tabularnewline 
Fast-MoCo (100 ep.)                           & \multicolumn{1}{c|}{73.6}  & \multicolumn{1}{c|}{73.5} & 72.5 \tabularnewline 
        \end{tabular}
	\end{center}
	\caption{Comparison of robustness to batch size from 1024 to 256. Result of the compared methods are from their original paper.}
	\label{R3}
\end{table}

\subsection{AdamW Optimizer}
When using AdamW as optimizer and pretrain for 100 epochs based on MoCo v3 ($69.3 \%$ w/ AdamW, result run by us) under the same setting, our Fast-MoCo can achieve 
$71.7 \%$ (+2.4\%) Top-1 accuracy for ImageNet linear evaluation, which demonstrates that our approach also works well with AdamW.

\subsection{Combine Stage Selection}
\begin{figure}[h]
    \centering
    \includegraphics[height=4cm]{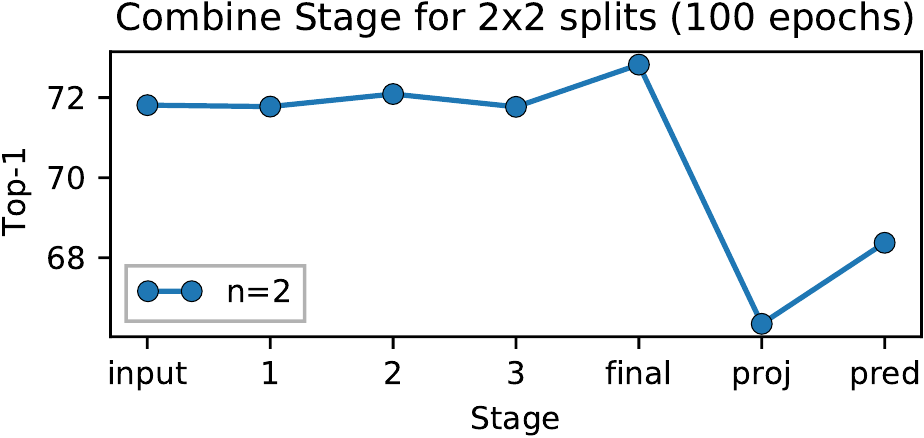}
    \caption{ImageNet linear evaluation accuracy (Y-axis) when Combine step is conducted at different stages (X-axis) for combining $n=2$ divided patches. Here ``input'' denotes the patches are combined at image level, $1 \sim 3$ denotes patches are combined at feature map level after ResNet stage $1 \sim 3$. ``final'', ``proj'' and ``pred'' denotes patches are combined at embedding level after encoder, projector, and predictor, respectively.}
    \label{fig:comb_ex}
\end{figure}

In Figure~\ref{fig:comb_ex} of this supplementary, we provide additional results on different choices of Combine stage. We have two target-sample pairs per image when number of patch combined $n=2$. When the combine stage is before ``final'', we stitch the feature map (for stage $1 \sim 3$) or patch image (for stage ``input'') vertically and horizontally with respect to their original position. Here we can see the performance drops significantly when the Combine step is conducted after the projection layer or prediction layer.

\section{Patch Encode Approaches}
\begin{figure}[h]
    \centering
    \includegraphics[height=13.8cm]{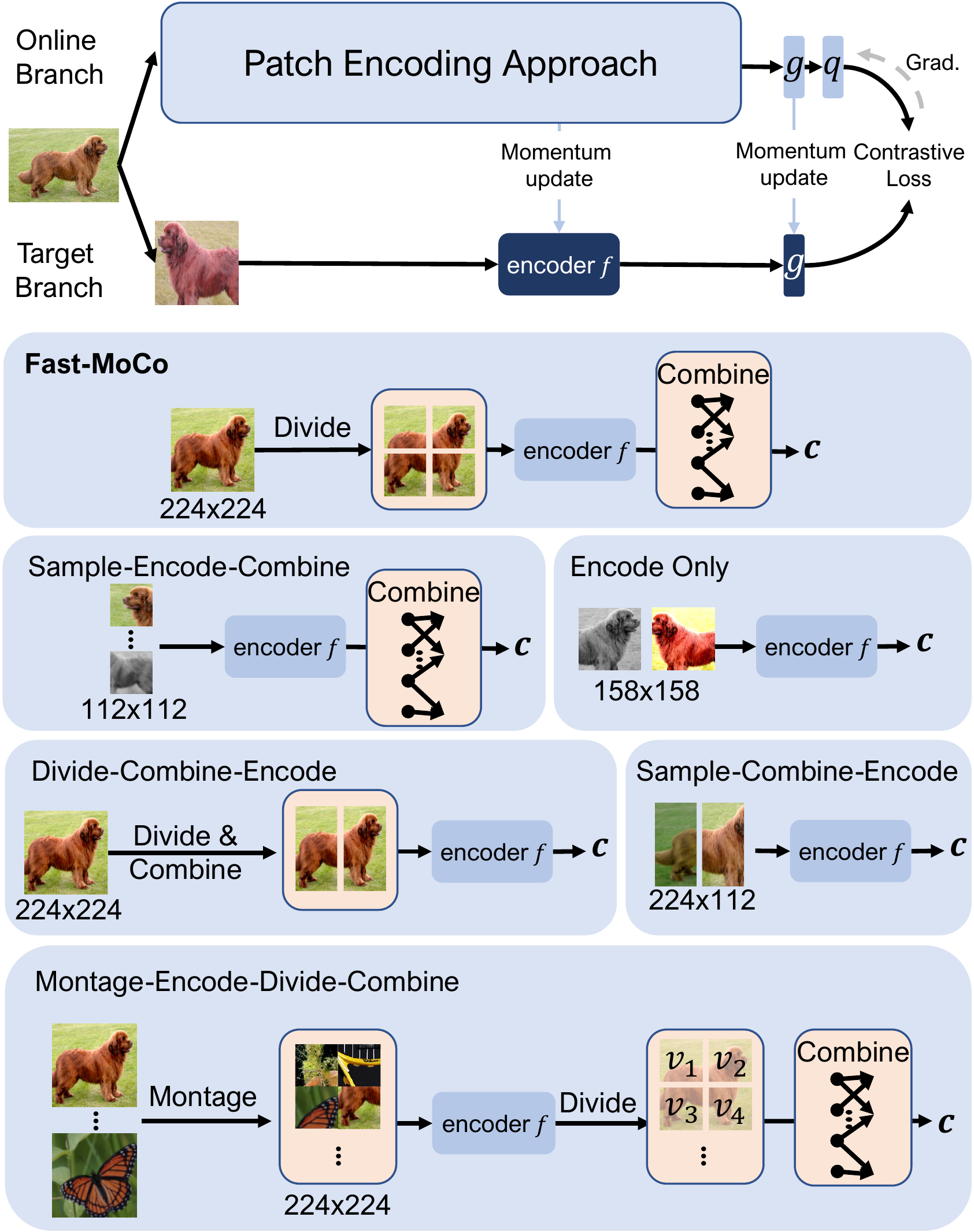}
    \caption{Illustration of patch encoding approaches compared in Section~\ref{sec:pea}}
    \label{fig:supp}
\end{figure}

In Figure~\ref{fig:supp}, we illustrate every patch encoding approach compared in Section~\ref{sec:pea}.

\end{document}